\pdfoutput=1
\documentclass[lettersize,journal]{IEEEtran}

\usepackage{amsmath,amssymb,amsfonts}
\usepackage{graphicx}
\usepackage{caption}
\usepackage{subcaption}
\usepackage{float}
\usepackage{makecell}
\usepackage{booktabs}
\usepackage{siunitx}
\usepackage{adjustbox} %
\usepackage{array}
\usepackage{multirow}
\usepackage{rotating}
\usepackage{algorithm}
\usepackage{algorithmic}

\usepackage{url}
\usepackage{xcolor}

\usepackage{etoolbox} %
\newif\ifarxiv
\arxivtrue %

\def\MYTITLE{Fast Event-based Optical Flow Estimation by Triplet Matching}

\usepackage[pagebackref=false,breaklinks=true,colorlinks,bookmarks=false]{hyperref}
\hypersetup{
  pdftitle={\MYTITLE},
  pdfsubject={Computer Vision, Robotics, Artificial Intelligence},
  pdfauthor={Shintaro Shiba, Yoshimitsu Aoki, Guillermo Gallego},
  pdfkeywords={Event Cameras, Intelligent Systems, Asynchronous Sensor, High Temporal Resolution}
}

\def\pol{p} %

\def\cE{\mathcal{E}} %
\def\numEvents{N_e} %
\def\numPixels{N_p} %
\def\numIter{N_\text{iter}}
\def\bx{\mathbf{x}}
\def\velflow{\mathbf{v}}

\def\cN{\mathcal{N}} %
\def\flow{\mathbf{f}}

\def\cH{\mathcal{H}} %

\def\textand{\;\text{and}\;}
\def\weight{w}
\def\triplet{T}

\usepackage[capitalize]{cleveref}
\crefname{section}{Sec.}{Secs.}
\Crefname{section}{Section}{Sections}
\Crefname{table}{Table}{Tables}
\crefname{table}{Tab.}{Tabs.}
\Crefname{figure}{Figure}{Figures}
\crefname{figure}{Fig.}{Figs.}

\usepackage{cite}

\newcommand{\novalue}{{\textendash}}

\ifarxiv
\newcommand{\gblue}[1]{#1} %
\else
\newcommand{\gblue}[1]{\textcolor{blue}{#1}} %
\fi

\definecolor{light-gray}{gray}{0.6}
\newcommand\gframe[1]{{\color{light-gray}\frame{#1}}}

\ifarxiv
\usepackage{balance}
\usepackage[absolute]{textpos}
\fi

\begin{document}
\title{\MYTITLE}

\ifarxiv
\definecolor{somegray}{gray}{0.6}
\newcommand{\darkgrayed}[1]{\textcolor{somegray}{#1}}
\begin{textblock}{11}(2.5, 0.4)
\begin{center}
\darkgrayed{This paper has been accepted for publication at the IEEE Signal Processing Letters, 2022.
\copyright IEEE}
\end{center}
\end{textblock}
\fi 

\author{Shintaro Shiba$^{1,2}$, Yoshimitsu Aoki$^{1}$, Guillermo Gallego$^{2,3}$
\thanks{$^1$ Department of Electronics and Electrical Engineering, Faculty of Science and Technology, Keio University, Kanagawa, Japan. 
$^2$ Department of Electrical Engineering and Computer Science, Technische Universit\"at Berlin, Berlin, Germany, 
$^3$ Einstein Center Digital Future and Science of Intelligence Excellence Cluster, Berlin, Germany.}%
}

\maketitle

\begin{abstract}
Event cameras are novel bio-inspired sensors that offer advantages over traditional cameras (low latency, high dynamic range, low power, etc.).
Optical flow estimation methods that work on packets of events trade off speed for accuracy, while event-by-event (incremental) methods have strong assumptions and have not been tested on common benchmarks that quantify progress in the field.
Towards applications on resource-constrained devices, it is important to develop optical flow algorithms that are fast, light-weight and accurate.
This work leverages insights from neuroscience, and proposes a novel optical flow estimation scheme based on triplet matching.
The experiments on publicly available benchmarks demonstrate its capability to handle complex scenes with comparable results as prior packet-based algorithms.
In addition, the proposed method achieves the fastest execution time ($>$ 10 kHz) on standard CPUs as it requires only three events in estimation.
We hope that our research opens the door to real-time, incremental motion estimation methods and applications in real-world scenarios.
\end{abstract}

\begin{IEEEkeywords}
Event cameras, optical flow, low latency, asynchronous sensor, neuroscience, robotics.
\end{IEEEkeywords}

\section{Introduction}

\IEEEPARstart{E}{vent} 
cameras \cite{Lichtsteiner08ssc,Finateu20isscc} have led to rethinking visual processing for various computer vision tasks because their operating principle and output data are fundamentally different from those of conventional, frame-based cameras.
These bio-inspired sensors naturally respond to the scene dynamics and offer advantages, such as low latency, high dynamic range (HDR) and data efficiency, which need to be unlocked with new algorithms \cite{Gallego20pami}.
Neuromorphic principles have been a major source of inspiration for such novel algorithms and hardware, especially in motion estimation tasks \cite{Haessig18tbcas,Brosch15fns,Paredes21neurips}.

Event-based optical flow estimation methods can be broadly classified as \emph{packet}-based or \emph{event-by-event}--based depending on how events are processed and update the estimator's output.
Packet-based methods process a batch of events (e.g., events in a fixed time window, say 10--100 ms, or a fixed number of events, typically 30k--1M), hence they require some waiting time before processing (inference) starts \cite{Gallego18cvpr,Benosman14tnnls,Shiba22eccv,Liu18bmvc}.
They trade off the high-speed advantages of event data for accuracy. 
Prior work has proposed adaptations of classical frame-based methods (block matching \cite{Liu18bmvc}, Lucas-Kanade \cite{Benosman12nn}),
spatio-temporal plane-fitting \cite{Benosman14tnnls,Akolkar22pami}, 
time-surface matching \cite{Nagata21sensors},
and contrast-maximization methods \cite{Zhu17icra,Gallego18cvpr,Shiba22eccv}.
While the above methods are model-based (optimization) methods, 
Artificial Neural Networks (ANN) \cite{Zhu18rss,Zhu19cvpr,Gehrig21threedv,Ziluo21arxiv,Lee20eccv} are also batch-based, 
and are inspired by frame-based ANN architectures \cite{teed2020eccv,jia2021ieeespl}, thus requiring data conversion into a tensor representation, such as voxel grids.
ANNs achieve current state-of-the-art accuracy for optical flow estimation and high speed \cite{Gehrig21threedv,Shiba22eccv}, but require power-hungry GPUs and lack interpretability.

\begin{figure}[t]
\centering
\includegraphics[clip,trim={5cm 0cm 3cm 2.5cm},width=.94\linewidth]{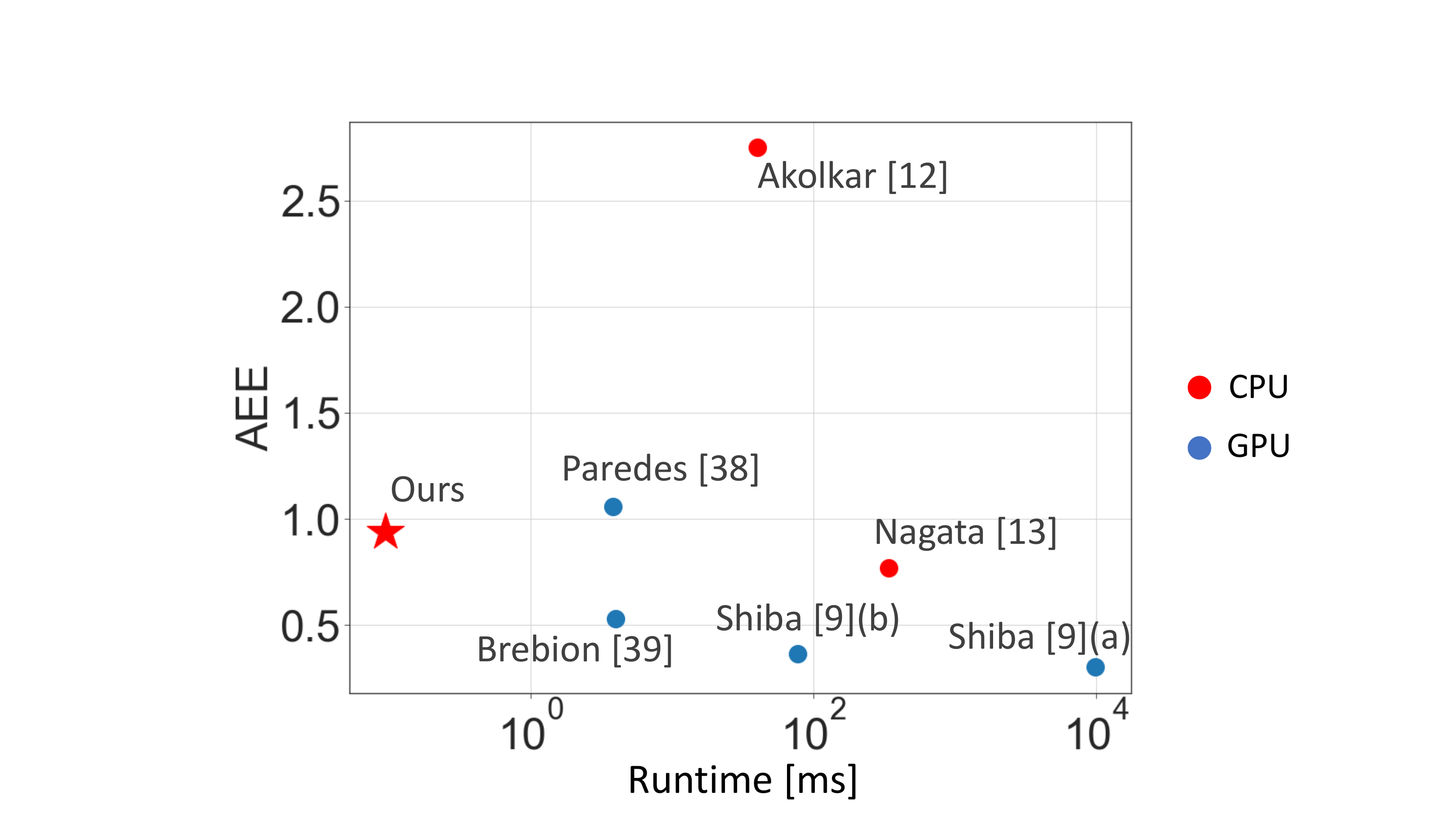}
\caption{\emph{Runtime vs.~accuracy comparison} for various event-based optical flow estimation methods. 
Results are on outdoor data of the MVSEC benchmark \cite{Zhu18rss} (see also \cref{tab:main_mvsec}). 
\gblue{Accuracy is measured based on Average Endpoint Error (AEE).}
Shiba \cite{Shiba22eccv}(a) and Shiba \cite{Shiba22eccv}(b) denote optimization-based and DNN-based results, respectively.
}
\label{fig:eyecatcher}
\end{figure}

\begin{figure*}[t]
\centering
{\includegraphics[clip,trim={0cm 11cm 0cm 0cm},width=\linewidth]{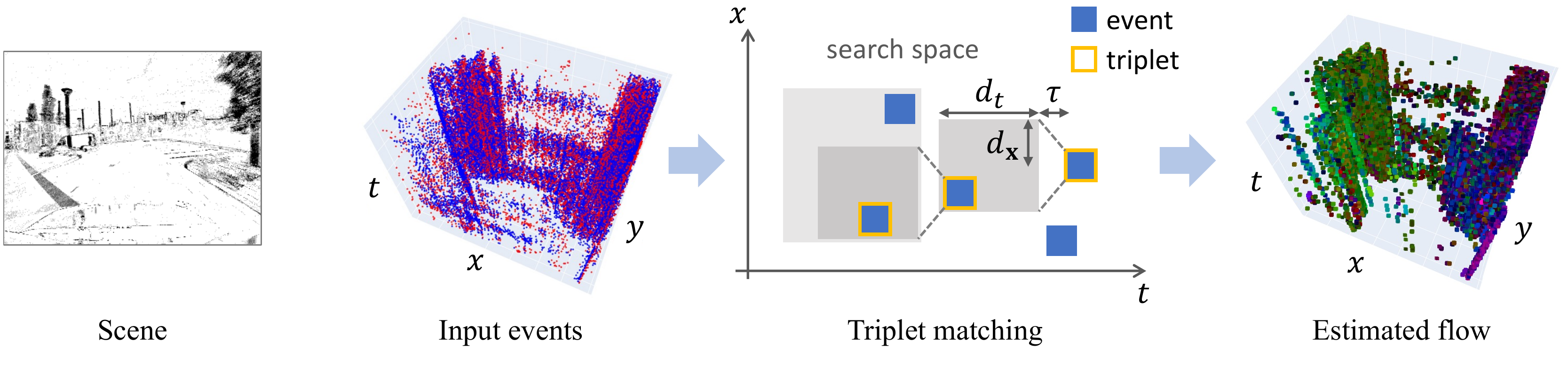}}
\caption{\emph{Triplet matchig algorithm}.
The algorithm seeks spatially and temporally neighboring events in an event-by-event manner,
and provides event-based flow $\flow_k$.
\gblue{For ease of visualization we only show the search in $x$ and $t$, but it is actually carried out in $x,y$ and $t$.}
Note this is an example of batch estimation given the input events.
}
\label{fig:method}
\end{figure*}

On the other hand, event-by-event methods
process every event incrementally as it occurs (without waiting time), aiming to leverage the camera's low-latency advantage \cite{Brosch15fns,DAngelo2020fns}.
Many event-by-event methods, such as Spiking Neural Networks (SNNs), are inspired by the brain (i.e., \emph{neuromorphic}), since the neural circuits of visual processing are thought to be event-driven.
While previous work propose SNN architectures \cite{Orchard13biocas,Paredes19pami,Paredes21neurips},
they comprise low-level physiological parameters of neurons (e.g., membrane potentials) that are difficult to interpret, validate and adjust to improve the estimation accuracy.
Indeed, insects and mammals have different low-level underlying mechanisms, while they have similar algorithmic steps to transform light into motion \cite{clark2016currentbio}.
Hence, it is important to find abstracted logical operations of motion estimation, rather than to mimic the entire physiological properties of neurons.
From a practical point of view, most event-by-event methods have been tested on simple scenes, as opposed to the more complex real-world scenes and publicly-available benchmarks of batch-based methods \cite{Rueckauer16fns}.
This may be attributed to the use of tailored hardware \cite{Haessig18tbcas}, strong assumptions of the scene, limited problem settings \cite{Brosch15fns} or the difficulty in defining event-by-event benchmarks on real data with \si{\micro\second} resolution.
Hence, it is important to explore event-by-event motion estimation algorithms that can solve complex, real-world problems.

This work leverages insights from neuroscience, especially from the classical Barlow-Levick model \cite{Barlow1965ud}, and proposes a novel optical flow estimation scheme based on correlation of occurrence.
In contrast to previous batch-based methods, it requires only three events (triplet) for estimation, which opens the door to future real-time incremental motion estimation methods.
Compared to previous event-by-event approaches, it is tested on publicly-available optical flow benchmarks to demonstrate its capability to handle real-world scenes with comparable results.
Additionally, it is based on logical operations, which enables a simple and efficient data structure implementation and execution on standard CPUs.
In summary, our contributions are twofold:
($i$) we present a novel event-by-event algorithm for optical flow estimation, theoretically derived from neuroscience insights,
and ($ii$) we practically demonstrate that it achieves comparable results as prior work while only requiring a CPU and being faster than optimization-based algorithms (\cref{fig:eyecatcher}). %

The signal processing in this work materializes the ideas in current neuroscience models, 
shedding light on what the strong and weak scenarios are, in order to improve the models.

\section{Methodology}
\label{sec:method}

\subsection{Event Camera}
\label{sec:method:eventcamera}

Event cameras acquire visual data in the form of asynchronous per-pixel brightness differences called ``events'' \cite{Lichtsteiner08ssc,Gallego20pami}.
An event $e_k \doteq (t_k, \bx_k, \pol_k)$ is triggered as soon as the logarithmic brightness at the pixel $\bx_k\doteq (x_k, y_k)^{\top}$ exceeds a preset threshold.
Here, $t_k$ is the timestamp of the event with \si{\micro\second} resolution,
and polarity $\pol_k \in \{+1,-1\}$ is the sign of the brightness change (i.e., increase vs.~decrease, respectively).

\subsection{Triplet Matching}
\label{sec:method:triplet}

The idea of the triplet matching comes from neuroscience models by Hassenstein-Reichardt \cite{Hassenstein1956or} and Barlow-Levick \cite{Barlow1965ud}.
These correlator models estimate motion by computing pairwise neural activities (e.g., spikes) in space and time \cite{clark2016currentbio}.
Especially, \cite{Fitzgerald2015elife} suggests that triplet correlations (the product of pairwise correlations for three spikes in space-time) improve motion estimation accuracy.
Here, we introduce the idea of the triplet-matching method as logical operations in space-time coordinates.
We build an incremental (event-by-event) estimation algorithm, and extend it into batch mode for testing because benchmarks are specified on a batch basis.

\subsubsection{Incremental Estimation}

It consists of two main steps: \emph{search} and \emph{update} (\cref{algo:increment}).
Events are split by polarity, following the idea of ON- and OFF- circuits in the brain \cite{clark2016currentbio}.
The search step finds triplets of events that are aligned (i.e., correlated) in space-time assuming a constant velocity model (\cref{fig:method}).
One of the events in the triplet is the incoming event, and the other two events are searched for within its space-time neighborhoods of size $d_{\bx}, d_t$.
The search has two steps: first the set of all potential 2nd events is determined; then the set of all potential 3rd events (compatible with the previous two in the triplet) refines the search.
In the update step, every triplet of events is characterized by a different velocity. 
The velocity (flow) $\flow_{k}$ for the incoming event $e_k$ is computed as the average of the velocities of all triplets.
Later, for benchmarking purposes, the flow is voxelized (quantized on a space-time grid) and smoothed.

\begin{algorithm}[t]
\caption{Triplet matching algorithm}
\label{algo:increment}
\begin{algorithmic}[1]
 \renewcommand{\algorithmicrequire}{\textbf{Input:}}
 \renewcommand{\algorithmicensure}{\textbf{Output:}}
 \REQUIRE $e_{k}$, $\cH^{k - 1}$ %
 \ENSURE $\cH^{k}$, $\flow_{k}$ \\
  \STATE Find event neighborhood $H_{k}$ in \eqref{eq:search}.
  \FOR {$i \in H_{k}$}
  \STATE Search for triplet candidates \eqref{eq:searchTwo}.
  \STATE Collect triplet $\triplet = (k, i, j)$
  \ENDFOR \\
  \STATE Calculate $\flow_{k} \leftarrow \{ \triplet \}$ in \eqref{eq:flow} \\
  \STATE Update $\cH^{k} \leftarrow \cH^{k-1}, H_{k}$
\end{algorithmic}
\end{algorithm}

In the \emph{search step}, since event data are sorted by timestamp $t$, we use index maps to make the search efficient, with complexity $O(\numEvents \log{\numEvents})$. 
The index map $H_{k}$ of an event $e_k$ consists of the indices of its space-time neighbors:
\begin{equation}
\label{eq:search}
H_{k} = \{i \mid t_{k} - \tau - d_t \leq t_i \leq t_{k} - \tau \textand \| \bx_k - \bx_i \|  \leq d_{\bx} \}.
\end{equation}
Parameters $d_t$ and $d_{\bx}$ decide the maximum admissible velocity of the flow,
and $\tau$ is a \emph{refractory period}, which limits the search space by assuming neighboring events in the moving edge do not exist at the same timestamp.
$d_t$ can also be interpreted as the \emph{delay} in the Barlow-Levick model.
For each new event $e_k$, we build a set of index maps $\cH^{k} = \{ H_{i} \}_{i=1}^{k}$ 
and output a set of event triplets $\{\triplet\} \doteq \{(e_k, e_i, e_j)\}$.
To find the triplet match we look for event indices $j$ that have roughly constant velocity with the event pairs $(e_k, e_i)$ where $i \in H_k$:
\begin{equation}
\label{eq:searchTwo}
J_{k,i} = \{j \in H_i \mid t_{i} - \tau - d_t \leq t_{j} \leq t_{i} - \tau \textand \bx_{i} - \bx_j = \bx_{k} - \bx_{i} \}.
\end{equation}

The \emph{update step} calculates the flow $\flow_{k}$ and updates the index map $\cH^{k}$.
$\cH^{k}$ is obtained by adding new $H_{k}$ to $\cH^{k-1}$ and removing old index maps (we keep the last 20000 index maps per polarity).
The flow $\flow_{k}$ is obtained as the weighted average
\begin{equation}
\label{eq:flow}
    \flow_k \doteq \frac{\sum_{T} \weight_T \velflow_T}{\sum_{T} \weight_T},
\end{equation}
where $\velflow_T \doteq (\bx_j - \bx_k)/(t_j - t_k)$ is the velocity of each triplet.
Since \eqref{eq:flow} gives accurate flow if the triplet is caused by the same scene edge,
we use the weight $\weight_T$ to estimate the probability that the triplet belongs to the same edge.
Assuming constant velocity, if $e_j$ is produced by the same edge that generates $e_k$ and $e_i$, the expected timestamp of $e_j$ is given by
$\hat{t}_j = t_i - \delta$, where $\delta \doteq t_k - t_i$.
Therefore, to account for errors in the timestamps between $\hat{t}_j$ and  $t_j$, 
we set the weight $\weight_T \doteq \cN(t_j;\hat{t}_j,\delta^2)$, 
where $\cN$ is the Gaussian density function. 
\gblue{The proposed average flow due to the triplets \eqref{eq:flow} may not necessarily equal the optical flow but this strong assumption is justified by the empirical results (\cref{sec:experim})}.

\subsubsection{Batch Estimation}

We extend the incremental (event-by-event) estimator to batch mode because current benchmarks are batch-based.
For a set of events $\cE \doteq \{e_k \}_{k=1}^{\numEvents}$, 
we create the index maps $\cH^{\numEvents}$ first, which takes $O(\numEvents^2 \log{\numEvents})$.
Then the flow is calculated looping over each event using \cref{algo:increment}.
The overall computational complexity is $O(\numEvents^2 \log{\numEvents})$.

For benchmarking with ground truth, the event-wise flow is converted into a voxel-wise flow, which also enhances space-time coherence.
We quantize the time coordinates of $\flow_k$ into bins, and take the average of the $\{\flow_k\}$ that lie in each voxel.
We also apply a non-zero average filter (take average of only non-zero values) with kernel size $3 \times 3$ for spatial smoothing.

The computational complexity of both approaches is summarized in \cref{tab:complexity}.
For comparison, we also report those of \gblue{state-of-the-art} methods: Contrast Maximization (CMax) \gblue{approaches \cite{Gallego18cvpr,Shiba22eccv}} and time-surface matching \cite{Nagata21sensors}.
The latter methods require additional complexity for the number of iterations $\numIter$. %
We report runtime comparisons in \cref{sec:experim:runtime}.

\sisetup{round-mode=places,round-precision=1}
\begin{table}[!h]
\centering
\caption{Complexity of algorithms, for batch estimation and event-by-event estimation.
}
\adjustbox{max width=\textwidth}{%
\setlength{\tabcolsep}{4pt}
\begin{tabular}{l*{3}{S}}
\toprule
& \text{Batch} & \text{Event-by-event}
\\
\midrule 
CMax \cite{Gallego18cvpr} & \text{$O(\numIter ( \numEvents + \numPixels))$} & \novalue %
\\[0.5ex]
Nagata et al. \cite{Nagata21sensors} & \text{$O(\numIter (\numEvents + \numPixels))$} & \novalue %
\\[0.5ex]
Ours & \text{$O(\numEvents^2 \log{\numEvents})$} & \text{$O(\numEvents\log{\numEvents})$} \\
\bottomrule
\end{tabular}
\label{tab:complexity}
}
\end{table}

\section{Experiments}
\label{sec:experim}

\subsection{Datasets and Evaluation Metrics}
\label{sec:experim:datasets}

The MVSEC dataset \cite{Zhu18ral} is a standard dataset for optical flow estimation \cite{Zhu19cvpr,Gehrig21threedv,Paredes21neurips,Shiba22eccv}.
The data consists of event camera, LiDAR, and camera poses.
The event camera (mDAVIS346 camera \cite{Taverni18tcsii}) provides events, grayscale frames and IMU data ($346 \times 260$ pix).
The ground truth optical flow is provided as the motion field from the camera velocity and the depth of the scene \cite{Zhu18rss}.
The sequences are indoors with a drone and outdoors with a car, and we evaluate on 63.5 million events spanning 265 seconds from both outdoor and indoor sequences.

We measure optical flow accuracy to evaluate our method. 
The metrics are the Average Endpoint Error (AEE) and the percentage of pixels with AEE greater than $3$ pixels (\% Out).
The time intervals for evaluation are $\Delta t=1$ grayscale frame (at $\approx$ 45Hz, i.e., 22.2ms) and $\Delta t=4$ frames (89ms).
Flow accuracy is evaluated only in pixels with valid ground truth.
All experiments use $d_{\bx} = \sqrt{2}$ pix, $d_{t} =100$ms and $\tau = 3$ms.

We also show \gblue{additional} results on the ECD dataset \cite{Mueggler17ijrr}, which is widely used for motion estimation \cite{Gallego17ral,Zhu17cvpr,Gu21iccv,Mueggler18tro}.
Each sequence provides events, frames, calibration, and IMU data (at 1 kHz) from a DAVIS240C ($240 \times 180$ pix) \cite{Brandli14ssc}, as well as ground truth camera poses (at 0.2 kHz).

\subsection{Optical Flow Estimation Accuracy}
\label{sec:experim:accuracy}

\begin{table}[!t]
\centering
\caption{Results on MVSEC dataset \cite{Zhu18rss}.
Methods are presented as unsupervised learning-based (USL) or model-based (MB).
For brevity, EV-FlowNet is abbreviated as EVFN. 
Nagata et al.~\cite{Nagata21sensors} evaluate on shorter time intervals; for comparison, we scale the errors to $\Delta t=1$.
}
\label{tab:main_mvsec}
\adjustbox{max width=\columnwidth}{%
\setlength{\tabcolsep}{2pt}
\begin{tabular}{ll*{6}{S[table-format=2.2]}}
\toprule
 & 
 & \multicolumn{2}{c}{outdoor\_day1}
 & \multicolumn{2}{c}{indoor\_flying1}
 & \multicolumn{2}{c}{indoor\_flying2}
 \\
 \cmidrule(l{1mm}r{1mm}){3-4}
 \cmidrule(l{1mm}r{1mm}){5-6}
 \cmidrule(l{1mm}r{1mm}){7-8}
\multicolumn{2}{c}{$\Delta t=1$} 
&\text{AEE $\downarrow$} & \text{\%Out $\downarrow$}
&\text{AEE $\downarrow$} & \text{\%Out $\downarrow$}
&\text{AEE $\downarrow$} & \text{\%Out $\downarrow$}\\
\midrule 

\multirow{5}{*}{\begin{turn}{90}
USL
\end{turn}}

 & EVFN 
 \cite{Zhu19cvpr} & 0.32 & 0.0 & 0.58 & 0.0 & 1.02 & 4.0 \\
 
 & EVFN (retrain) \cite{Paredes21cvpr}  & 0.92 & 5.4 & 0.79 & 1.2 & 1.40 & 10.9\\
 
 & FireFlowNet \cite{Paredes21cvpr} & 1.06 & 6.6 & 0.97 & 2.6 & 1.67 & 15.3 \\
 & ConvGRU-EVFN \cite{Paredes21neurips}  & 0.47 & 0.25 & 0.60 & 0.51 & 1.17 & 8.06 \\

 & MultiCM-EVFN \cite{Shiba22eccv}  & 0.3632 & 0.087 & \novalue & \novalue & \novalue & \novalue\\
 \midrule 

\multirow{5}{*}{\begin{turn}{90}
MB
\end{turn}}
 & Nagata et al. \cite{Nagata21sensors} & 0.7696 & \novalue & 0.62 & \novalue & 0.9324 & \novalue \\
 & Akolkar et al. \cite{Akolkar22pami} & 2.75 &  \novalue & 1.52 & \novalue & 1.59 & \novalue \\
 & Brebion et al. \cite{Brebion21tits} & 0.53 & 0.2 & 0.52 & 0.1 & 0.98 & 5.5 \\
 & MultiCM \cite{Shiba22eccv}  & 0.30075 & 0.10440 & 0.42274 & 0.10303 & 0.60456 & 0.59485 \\
 & Ours & 0.9384639346 & 3.07899511 & 1.0526714583260 & 2.901922235 & 1.6768836 & 13.435257 \\

\midrule 
 \\[-0.2ex]
\multicolumn{2}{c}{$\Delta t=4$}\\
\midrule
\multirow{3}{*}{\begin{turn}{90}
USL
\end{turn}}
 
 & EVFN \cite{Zhu19cvpr}& 1.30 & 9.7 & 2.18 & 24.2 & 3.85 & 46.8 \\
 & ConvGRU-EVFN \cite{Paredes21neurips} & 1.69 & 12.50 & 2.16 & 21.51 & 3.90 & 40.72 \\
 & MultiCM-EVFN \cite{Shiba22eccv}  & 1.488 & 11.72 & \novalue & \novalue & \novalue & \novalue \\

 \midrule 

\multirow{2}{*}{\begin{turn}{90}
MB
\end{turn}}
 & MultiCM \cite{Shiba22eccv}  & 1.24994 &  9.20702 & 1.68961 & 12.94904 & 2.48956 & 26.34562 \\
 & Ours & 3.5993733109102086 & 49.04043777877478 & 4.059380506703 & 53.884642296996965 & 6.391457718480932 & 71.82462229105026 \\

\bottomrule
\end{tabular}
}
\end{table}

\Cref{tab:main_mvsec} comprises flow estimation accuracy results on the MVSEC benchmark.
The top part of the table reports results for $\Delta t=1$, and the bottom part reports $\Delta t=4$.
The methods in the table are categorized as unsupervised learning-based (USL), i.e., using a Deep Neural Network (DNN) on grid-converted events, and model-based (MB).
Results for $\Delta t=1$ are thorough, with our method in the middle accuracy range among all methods.
Results for $\Delta t=4$ are not as complete because the literature does not report them (especially most model-based methods). 
While a thorough comparison for $\Delta t=4$ is difficult, our error is roughly four times bigger than for $\Delta t=1$, which makes sense, 
and it is consistently 2.5--3 times bigger than that of the most accurate method \cite{Shiba22eccv} for both $\Delta t=\{1,4\}$.
The fact that for longer time intervals batch-based methods ($\Delta t=4$) achieve higher accuracy than our method may be attributed to the fact that our method is event-by-event, so it does not leverage long-term temporal smoothing,
which would improve robustness to noise.

\def\figWidth{0.43\linewidth}
\begin{figure}[t]
\centering
\begin{subfigure}{\figWidth}
  \centering
  \gframe{\includegraphics[width=\linewidth]{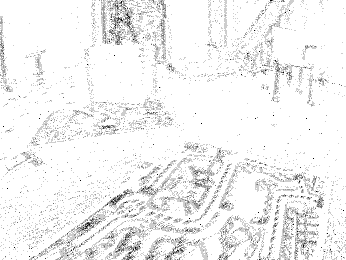}}
  \caption{Input events\label{fig:mvsec:inputevents}}
\end{subfigure}\;
\begin{subfigure}{\figWidth}
  \centering
  \gframe{\includegraphics[width=\linewidth]{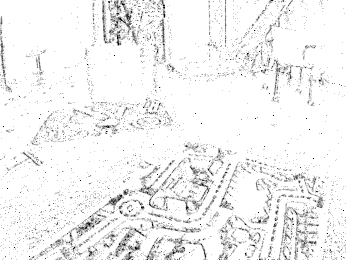}}
  \caption{IWE\label{fig:mvsec:iwe}}
\end{subfigure}\\[1ex]
\begin{subfigure}{\figWidth}
  \centering
  \gframe{\includegraphics[width=\linewidth]{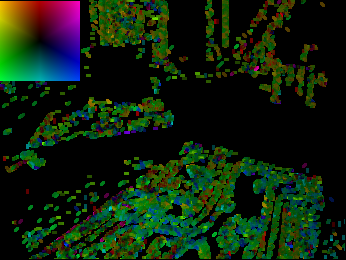}}
  \caption{Estimated flow\label{fig:mvsec:estimatedflow}}
\end{subfigure}\;
\begin{subfigure}{\figWidth}
  \centering
  \gframe{\includegraphics[width=\linewidth]{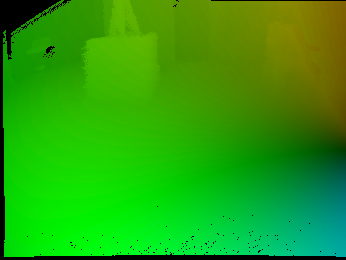}}
  \caption{Ground truth flow\label{fig:mvsec:gt}}
\end{subfigure}\\
\caption{\gblue{\emph{Optical flow results on MVSEC data.}}
}
\label{fig:mvsec}
\end{figure}

\gblue{The results in \cref{fig:mvsec} show that} the events displaced using the estimated flow produce sharp images of warped events (\gblue{\cref{fig:mvsec:iwe}}, IWEs \cite{Gallego18cvpr}).
The Flow Warp Loss \cite{Stoffregen20eccv} measures the sharpness of the IWE: $1.154$ for outdoor\_day1, $1.157$ for indoor\_flying1, and $1.248$ for indoor\_flying2, where $\text{FWL}>1$ indicates sharper than the identity warp baseline (i.e., zero flow).
The figure also shows the estimated flow (\gblue{\cref{fig:mvsec:estimatedflow}, and color wheel}); notice that our method produces a flow vector for each event (\cref{fig:method}), whereas it is common to display the flow for every pixel (image-based legacy).
Hence, \cref{fig:mvsec} shows a 2D collapsed version of the estimated space-time optical flow field, for visual comparison with the ground truth (\gblue{\cref{fig:mvsec:gt}}).
The flow is most reliably estimated in regions where events happen, i.e., scene edges.
Further spatial and temporal smoothness could be enhanced if needed:
for example, homogeneous brightness regions between edges could be filled in by some prior, such as a regularizer or in-painting algorithm.

\subsection{Runtime Comparison}
\label{sec:experim:runtime}

The proposed method runs in an event-by-event manner, hence trades off accuracy for speed, compared with batch-based methods.
We showed computational complexity comparison in \cref{tab:complexity}.
Now, we conduct the runtime comparison among several previous work.
We use Python (3.9.12) on CPUs (Mac M1 2020, 8 Cores) and average the runtime of processing 300k events incrementally.
The results are shown in \cref{fig:eyecatcher}.
Our method achieves the fastest runtime among compared methods: 0.0934 milliseconds ($>$10 kHz).
Note that many methods in the literature, \gblue{such as the 2nd and 3rd fastest ones \cite{Paredes21cvpr,Brebion21tits}, use GPUs, while ours runs natively on CPUs.
This is crucial for applications on resource-constrained platforms.}

\subsection{Effect of Pixel Quantization}
A limitation of the proposed method is the quantization of the flow direction since the search for the second event in the triplet is limited to the 8 neighboring pixels of the current event.
To illustrate it, we conduct experiments on the \emph{dynamic\_translation} sequence from the ECD dataset \cite{Mueggler17ijrr}.
\Cref{fig:quantization} shows the distribution of $\velflow_T$ over all events (assuming a planar translation model, i.e., constant velocity over all pixels).
Similar to the SNN proposed in \cite{Orchard13biocas}, $\velflow_T$ is constrained to eight cardinal directions.
However, in contrast to \cite{Orchard13biocas}, which quantizes both the direction and magnitude of the flow, our method can estimate a continuum of magnitudes.
The distributions are spread around a main direction and its two neighboring ones, which is due to the small aperture ($5 \times 5$ pix) used for each triplet.

\def\figWidth{0.445\linewidth}
\def\figWidthLong{0.53\linewidth}
\begin{figure}[t]
\centering
\begin{subfigure}{\figWidth}
  \centering
  \gframe{\includegraphics[width=\linewidth]{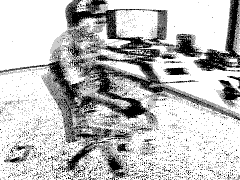}}
\end{subfigure}\;
\begin{subfigure}{\figWidthLong}
  \centering
  {\includegraphics[clip,trim={0.8cm 0.8cm 0.5cm 0.8cm},width=\linewidth]{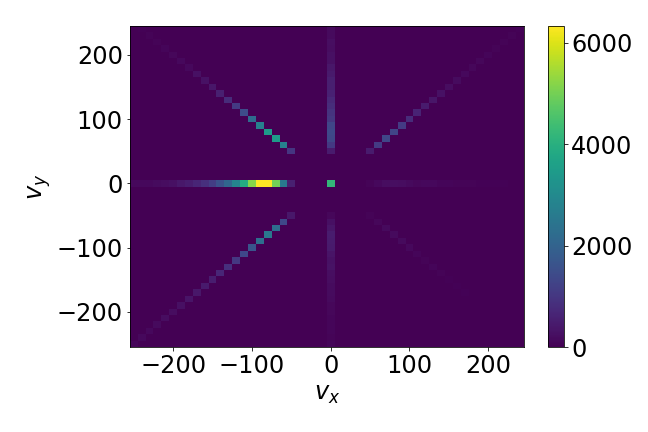}}
\end{subfigure}\\[2ex]
\begin{subfigure}{\figWidth}
  \centering
  \gframe{\includegraphics[width=\linewidth]{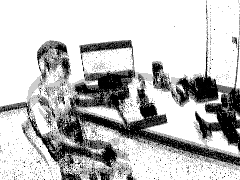}}
  \caption{Input events}
\end{subfigure}\;
\begin{subfigure}{\figWidthLong}
  \centering
  {\includegraphics[clip,trim={0.8cm 0.8cm 0.5cm 0.8cm},width=\linewidth]{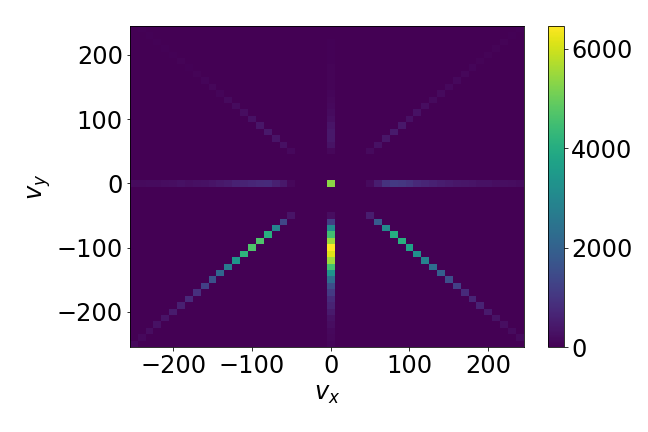}}
  \caption{Estimated velocities $\velflow_T$}
\end{subfigure}\\
\caption{\emph{Effect of pixel quantization
\gblue{on ECD data}}.
In the top row the motion is dominantly horizontal, whereas in the bottom row it is vertical, as can be seen by the thickness of the edges (left) and the velocity distributions (right).}
\label{fig:quantization}
\end{figure}

\section{Conclusion}
\label{sec:conclusion}

We proposed a novel event-based optical flow estimation scheme based on triplet matching inspired by motion estimation models in neuroscience.
The experiments demonstrate that it is considerably fast ($>$ 10 kHz) on standard CPUs while providing comparable results as prior batch-based algorithms.
We hope that our work opens the door to real-time, realistic, incremental motion estimation methods and event-camera applications on resource-constrained devices.

\section{Acknowledgment} 
We thank Dr. R. Tanaka for useful discussions.
This research was funded by the German Academic Exchange Service (DAAD), Research Grant-Bi-nationally Supervised Doctoral Degrees/Cotutelle, 2021/22 (57552338) and the Deutsche Forschungsgemeinschaft (DFG, German Research Foundation) under Germany’s Excellence Strategy – EXC 2002/1 ``Science of Intelligence'' – project number 390523135.

\ifarxiv
\balance
\fi
\bibliographystyle{IEEEtran}

\end{document}